\documentclass[10pt,twocolumn]{article}

\usepackage[T1]{fontenc}
\usepackage[utf8]{inputenc}
\usepackage{microtype}
\usepackage[margin=0.75in]{geometry}
\usepackage{amsmath,mathtools}
\usepackage{newtxtext,newtxmath}
\usepackage{booktabs}
\usepackage{graphicx}
\usepackage{subcaption}
\usepackage{enumitem}
\usepackage{xcolor}
\usepackage{xspace}
\usepackage{array}
\usepackage{placeins}
\usepackage[super,sort&compress]{natbib}
\usepackage{url}
\definecolor{linkblue}{RGB}{20,72,140}
\usepackage[
  colorlinks=true,
  allcolors=linkblue,
  breaklinks=true,
  bookmarksnumbered=true
]{hyperref}
\usepackage[nameinlink,capitalise,noabbrev]{cleveref}

\newcounter{finding}
\newcommand{\finding}[1]{%
  \refstepcounter{finding}%
  \par\noindent
  \begingroup
  \setlength{\fboxsep}{0pt}%
  \colorbox{black!6}{%
    \hbox{%
      {\color{black!45}\vrule width 2.2pt}%
      \hspace{4.5pt}%
      \parbox{\dimexpr\columnwidth-2.2pt-9pt\relax}{%
        \vspace*{2.8pt}%
        \textbf{Finding \thefinding.} #1\par
        \vspace*{2.8pt}%
      }%
      \hspace{4.5pt}%
    }%
  }%
  \endgroup
  \par
}

\newcommand{\papertitle}{CacheReforge: Bounded Recovery for Stale KV Caches under Evolving Adapters}

\title{\parbox{0.96\textwidth}{\centering
{\fontsize{22}{24}\selectfont\bfseries CacheReforge:}\\[0.16em]
\resizebox{0.94\textwidth}{!}{\bfseries Bounded Recovery for Stale KV Caches under Evolving Adapters}}}
\author{%
  \begin{tabular}{cc}
    Yuhang Cao\textsuperscript{\ensuremath{\dagger}} &
    Yanzhou Mu\textsuperscript{\ensuremath{\dagger}}\\[0.02em]
    {\footnotesize\texttt{caoyuhang@smail.nju.edu.cn}} &
    {\footnotesize\texttt{602022320006@smail.nju.edu.cn}}\\[0.4em]
    Chunrong Fang\textsuperscript{*} &
    Zhenyu Chen\textsuperscript{*}\\[0.02em]
    {\footnotesize\texttt{fangchunrong@nju.edu.cn}} &
    {\footnotesize\texttt{zychen@nju.edu.cn}}
  \end{tabular}\\[2.4em]
  Nanjing University\\[-0.1em]
  {\footnotesize\textsuperscript{\ensuremath{\dagger}}Equal contribution\quad
   \textsuperscript{*}Corresponding authors}
}
\date{}

\hypersetup{%
  pdftitle={\papertitle},
  pdfauthor={Yuhang Cao, Yanzhou Mu, Chunrong Fang, Zhenyu Chen},
  pdfsubject={KV cache recovery under continually evolving adapters},
  pdfkeywords={KV cache, LoRA, adapters, LLM inference, cache recovery, model serving}
}

\begin{document}

\maketitle

\begin{abstract}

Large language models rely on KV caching to reduce repeated prefill computation in long context and interactive applications. As lightweight adapters evolve, cached states reflect earlier versions, so stale reuse distorts current model outputs, while complete affected suffix recomputation restores fidelity at substantial cost. We seek minimal recomputation that recovers current adapter behavior. Existing systems track token, context, or stable adapter identity, but neither represent caches from earlier adapter versions nor distinguish update propagation from the recomputation required for behavioral recovery.
To address these gaps, we introduce \textsc{CacheReforge}, which represents stale KV caches as layerwise mixed-version objects. It combines per-layer adapter anchors, calibrated sensitivity, accumulated drift, and executable restart boundaries to select direct reuse, bounded recomputation, or complete affected-suffix recovery. We distinguish dependency depth from the functional recomputation horizon and use cumulative tail influence to characterize when bounded recovery preserves current-model behavior.
We evaluate \textsc{CacheReforge} on Qwen2.5-1.5B and Qwen2.5-7B with continual LoRA updates, including 16K HotpotQA and 2WikiMQA workloads. \textsc{CacheReforge} reduces mean KL divergence by $92.4\%$ relative to stale reuse, while recomputing only $5.44\%$ of layers and reducing cache-maintenance time by $93.2\%$ relative to fresh full prefill. These results show that version-aware recovery preserves model fidelity and most KV caching gains.\looseness=-1

\end{abstract}

\section{Introduction}

Large language models (LLMs) are increasingly deployed in real-world applications, such as knowledge-intensive question answering and code generation \citep{ouyang2022training,lewis2020retrieval,chen2021evaluating}. These applications expose models to evolving knowledge, user preferences, and interaction patterns, creating a growing need for continual adaptation after deployment. Parameter-efficient methods such as LoRA make this process practical by freezing the base model and updating a lightweight adapter \citep{hu2022lora}. In LLM test-time training and related online adaptation settings, incoming data can trigger successive updates to the same adapter \citep{sun2020ttt}. We call this process \emph{adapter evolution}.

However, continual adaptation conflicts with efficient long-context inference. KV caching reduces repeated prefix computation by reusing stored key and value states \citep{vaswani2017attention,pope2023efficiently,kwon2023vllm}, and systems such as Prompt Cache and SGLang build on this mechanism \citep{gim2024promptcache,zheng2024sglang}. These systems assume consistent model parameters across cache creation and reuse, but adapter evolution breaks this assumption. An earlier adapter version may produce a \emph{stale KV cache} that no longer matches current-model inference, even for the same prefix. Direct reuse preserves computation but may distort outputs, while recomputing the complete affected suffix restores fidelity but reduces caching gains. The resulting challenge is to recover current adapter behavior with minimal recomputation despite condition-specific invalidation and downstream dependencies.

Solving this cache recovery problem requires version-aware validity tracking and fine-grained recomputation, yet existing research leaves two key gaps. \textbf{Version-blind validity.} Current KV caching and adapter-serving systems judge cache validity from token and context identity or stable adapter identities \citep{gim2024promptcache,zheng2024sglang,yao2025cacheblend}. They cannot represent a cache produced by an earlier version of the same evolving adapter, even when the token prefix remains unchanged. \textbf{Coarse recovery.} Existing policies either reuse the stale cache or recompute the complete affected suffix. They therefore cannot distinguish how far an update propagates from how far recomputation must proceed to restore current-model behavior.

To close these gaps, we introduce \textsc{CacheReforge}, a version-aware recovery framework for continually evolving adapters. It represents a partially refreshed KV cache as a layerwise mixed-version object and records the producing adapter state for each layer, enabling precise validity tracking across successive updates. After each update, \textsc{CacheReforge} locates the earliest executable restart layer, combines calibrated layer sensitivity with accumulated adapter drift, and selects direct reuse, bounded recomputation, or complete affected-suffix recovery. It refreshes only the required contiguous range, updates the corresponding layer anchors, and preserves the historical identities of unrecomputed states. We further distinguish dependency depth from the functional recomputation horizon and use cumulative tail influence to derive KL and top-1 guarantees and guide the runtime risk proxy. This design jointly provides version-aware cache representation and adaptive recovery granularity.

We evaluate \textsc{CacheReforge} through controlled and realistic experiments on Qwen2.5-1.5B and Qwen2.5-7B. We freeze the base model, continually update a rank-8 LoRA adapter, and use 4K and 16K controlled workloads together with 16K HotpotQA and 2WikiMQA evaluations.

\noindent\(\bullet\)
\textbf{RQ1: Does adapter evolution make stale KV caches unsafe?} We compare stale reuse with fresh recomputation across adapter versions, update targets, and layer positions. Output divergence and losses in task quality generally increase with the version gap, and stale reuse can erase most gains from adaptation. These results establish stale-cache recovery as a necessary systems problem.

\noindent\(\bullet\)
\textbf{RQ2: How much recomputation does recovery require?} We evaluate candidate recovery windows across $72$ update profiles and identify the first window that satisfies distributional, prediction, and task-level quality criteria. Direct reuse succeeds in $37$ profiles, bounded recovery succeeds in $18$, and $17$ require the complete affected suffix. Recovery therefore depends on the update condition, and additional recomputation does not always improve fidelity.

\noindent\(\bullet\)
\textbf{RQ3: What determines the recovery boundary?} We analyze how update-induced errors propagate and compare cumulative stale-suffix signals with boundary-only measurements. Cumulative suffix error correlates more strongly with window KL, reaching Spearman correlations of $0.873$ overall and $0.690$ within conditions. This result supports cumulative tail influence as the basis for layerwise recovery risk.

\noindent\(\bullet\)
\textbf{RQ4: Does drift-sensitive recovery improve efficiency and fidelity?} We evaluate consecutive adapter updates over $96$ held-out trajectories and compare \textsc{CacheReforge} with fresh full recomputation, exact affected-suffix recovery, stale reuse, and two policy ablations. \textsc{CacheReforge} reduces mean KL by $92.4\%$ relative to stale reuse while recomputing $5.44\%$ of layers and reducing cache-maintenance time by $93.2\%$ relative to fresh full prefill. These results show that layerwise drift sensitivity and adaptive range selection recover most current-model fidelity at substantially lower cost.\looseness=1

Overall, our work makes three main contributions:

\begin{itemize}
    \item \textbf{Problem Formulation and Theory.}
    We formulate stale KV cache recovery under evolving adapters and represent partially refreshed caches as layerwise mixed-version states. We distinguish dependency depth from the functional recomputation horizon, define cumulative tail influence, and derive sufficient conditions for next-token KL and top-1 consistency.

    \item \textbf{Version-Aware Recovery.}
    We develop \textsc{CacheReforge}, which combines module-aware restart points, per-layer adapter anchors, calibrated sensitivity, and accumulated drift to select direct reuse, bounded recomputation, or complete affected-suffix recovery.

    \item \textbf{Empirical Evaluation.}
    We evaluate Qwen2.5 models across update conditions and continual adapter trajectories. The results reveal three recovery outcomes, validate cumulative suffix influence as a stronger predictor than boundary error, and show that \textsc{CacheReforge} preserves model fidelity with substantially lower recomputation and maintenance cost.
\end{itemize}


\section{Background and Problem Formulation}

\begin{figure}[t]
    \centering
    \includegraphics[width=\columnwidth]{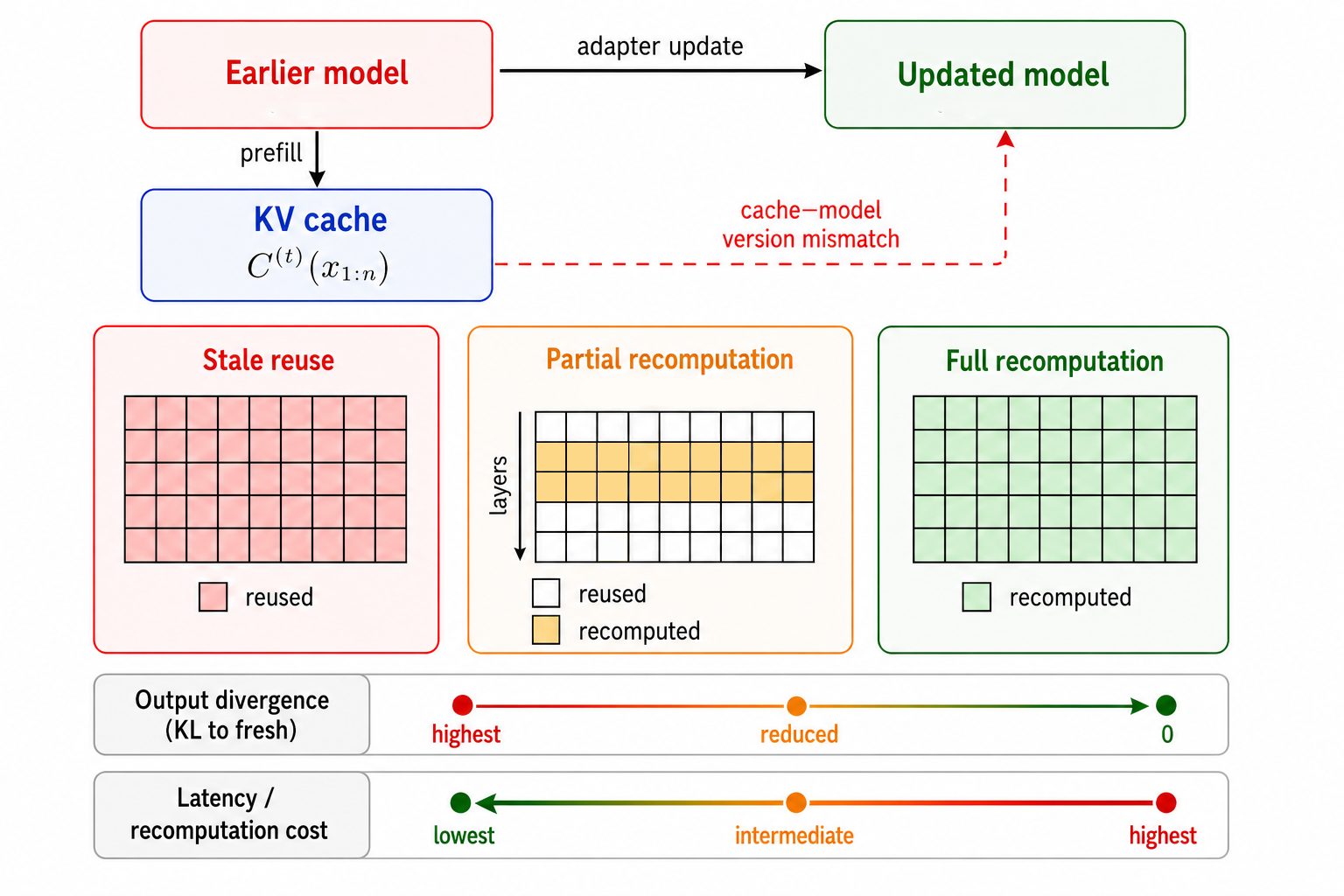}
    \caption{Overview of Cache Recovery under Adapter Evolution.\looseness=-1}
    \label{fig:cache-recovery-tradeoff}
\end{figure}

In this section, we first introduce KV caching under adapter evolution, explaining why a KV cache for an unchanged prefix becomes version-dependent when the same inference-time adapter evolves: although the tokens remain fixed, the stored states were produced by an earlier model. We then formalize the cache recovery problem, defining recovery as finding the shortest executable layer range that restores behavior relative to fresh prefill under the current adapter. This formulation distinguishes full-prefill, exact affected-suffix, and bounded recomputation, and exposes the central question of whether the complete affected suffix is functionally necessary. Figure~\ref{fig:cache-recovery-tradeoff} previews KV caching under adapter evolution and the cache recovery problem.\looseness=-1

\subsection{KV Caching under Adapter Evolution}

KV caching stores the key and value states produced by Transformer self-attention during prefix processing and reuses them during subsequent decoding, avoiding repeated prefill computation \citep{vaswani2017attention,pope2023efficiently,kwon2023vllm}. This mechanism is particularly valuable for long contexts and repeated prefixes, including shared system prompts, multi-turn conversations, and repeated generation from the same context. Prompt Cache organizes reusable prefix states into modular components, while SGLang reuses shared prefixes across structured generation calls \citep{gim2024promptcache,zheng2024sglang}.

Under adapter evolution, a system may retain a long-prefix KV cache across successive adapter updates to avoid processing the same prefix again. However, the retained cache was generated by an earlier adapter version, whereas current inference uses the updated adapter. The cache is therefore no longer guaranteed to represent the current model state.

\subsection{Cache Recovery Problem}

The preceding setting leaves the retained cache tied to an earlier adapter version. We now formalize when such a cache becomes stale and what its recovery must achieve. Consider a prefix $x_{1:n}$. Prefill by a Transformer with $N$ layers indexed
$0,\ldots,N-1$ under adapter version $v_t$ produces the layerwise KV cache
\begin{equation}
    \mathcal{C}^{(t)}(x_{1:n})
    = \left\{K_{\ell}^{(t)},V_{\ell}^{(t)}\right\}_{\ell=0}^{N-1}.
    \label{eq:versioned-cache}
\end{equation}
The superscript emphasizes that cache validity depends on both the token prefix and the adapter version that produced it. If the current adapter has advanced to $v_{t+\Delta}$ while the stored cache remains $\mathcal{C}^{(t)}$, we call it a \emph{stale KV cache} with version gap $\Delta$. Reusing this cache avoids prefill computation, but the resulting inference may deviate from fresh recomputation under $v_{t+\Delta}$, which serves as our reference behavior.

Directly correcting historical K/V states from the adapter update is generally insufficient because downstream states also depend on update-altered hidden representations. Accurately reconstructing these effects would largely repeat the affected forward computation and erase the intended computational advantage, so we focus on recomputation-based recovery.

The cache-recovery problem is therefore to determine the minimum executable recomputation needed to recover current-model behavior.

Throughout, \emph{fresh full-prefill recomputation} reruns all $N$ layers,
\emph{exact affected-suffix recomputation} reruns from the executable restart
layer through layer $N-1$, and \emph{bounded recovery} stops before that endpoint.
The remaining question is whether recovering fresh-model behavior always requires the complete affected suffix or can be achieved with a shorter executable range.


\section{Theoretical Analysis}

This section asks how much of an affected cache must be recomputed
after an adapter update.  Exact dependency alone cannot answer this question:
an update may reach every downstream state even when the remaining stale suffix
has little output effect.  We therefore define \emph{tail influence} to measure
this residual effect and separate dependency depth from the recomputation
horizon.  We then derive KL and top-1 guarantees from the same quantity.  These
offline guarantees establish when bounded recovery can preserve fresh-model
behavior and motivate the empirical runtime policy developed next.
\looseness=-1

\subsection{Tail Influence}

We estimate tail influence through an offline sequence of hybrid executions.  Let
$z^{\star}$ be the first-step next-token logits from fresh recomputation under
the current adapter.  For a boundary $b$, let $z^{(b)}$ use refreshed cache
states before layer $b$ and historical states at layers $b,\ldots,N-1$.
The endpoint $z^{(N)}=z^{\star}$ replaces the entire stale suffix.

Moving the boundary from $\ell$ to $\ell+1$ isolates the output change caused by
refreshing one additional cache layer.  We denote this incremental influence by
$v_{\ell}$ and define the cumulative \emph{tail influence} after boundary $b$ as
\begin{equation}
    v_{\ell}=z^{(\ell)}-z^{(\ell+1)},
    \qquad
    T_b=\sum_{\ell=b}^{N-1}\lVert v_{\ell}\rVert_2.
    \label{eq:tail-influence}
\end{equation}
$T_b$ measures the total unresolved influence of the cache states that remain
stale.  It is intentionally cumulative: a small error at the restart boundary
does not guarantee that later stale layers are harmless, while several
individually visible effects may partially cancel at the output.  Because $T_b$
sums influence magnitudes, it forms a monotone conservative envelope: it does
not itself identify local reversals, which depend on the directions of the
underlying $v_{\ell}$.  The quantity also does not assume that perturbations
shrink monotonically through the network.  It only asks whether
the remaining stale suffix has enough aggregate output influence to matter.

This view separates two notions that are otherwise easy to conflate.  The
\emph{dependency depth} describes how far an update can propagate in the exact
computation graph.  The \emph{recomputation horizon} describes how far recomputation
must proceed before the unresolved tail becomes functionally negligible.  For
condition $c$ with restart $r(c)$, let $p_{c,h}$ and $p_c^{\star}$ denote the
distributions after recomputing $h$ layers and under fresh recomputation.  For
discrepancy $\mathcal D$ and tolerance $\epsilon$, define
\begin{equation}
    H_{\epsilon}(c)=\min_{0\leq h\leq N-r(c)}
    \{h:\mathbb{E}_x[\mathcal D(p_c^{\star},p_{c,h})]\leq\epsilon\}.
    \label{eq:recomputation-horizon}
\end{equation}
This first-hit definition does not assume that recovery error decreases
monotonically with $h$; a larger window may still exhibit a local reversal.
Thus $H_{\epsilon}(c)$ may be zero, bounded before the suffix endpoint, or equal
to the complete executable suffix.  Dependency depth may cover the complete
downstream network even when this horizon is much shorter.

\subsection{Functional Guarantees}

We next translate tail influence into sufficient conditions for distributional
and prediction consistency.  Let $p^{(b)}=\operatorname{softmax}(z^{(b)})$ and
$p^{\star}=\operatorname{softmax}(z^{\star})$ be the corresponding next-token
distributions.  The hybrid logits admit a telescoping decomposition, and the
smoothness of log-sum-exp converts the remaining logit influence into a
conservative distributional guarantee:
\begin{equation}
    D_{\mathrm{KL}}\!\left(p^{\star}\Vert p^{(b)}\right)
    \leq \frac{1}{4}T_b^2.
    \label{eq:tail-kl-certificate}
\end{equation}
The bound directly characterizes the recomputation horizon.  For a KL tolerance
$\epsilon$, any executable recovery boundary with $T_b\leq 2\sqrt{\epsilon}$
guarantees next-token KL within that tolerance.  This sufficient condition does
not imply monotonic observed KL, because layerwise effects may reinforce or
cancel.

Distributional similarity does not by itself state whether the next-token
prediction is unchanged.  Let $\gamma^{\star}$ be the margin between the largest and
second-largest fresh logits.  The fresh top-1 prediction is preserved whenever
the remaining tail is smaller than half of this margin:
\begin{equation}
    T_b<\frac{\gamma^{\star}}{2}
    \quad\Longrightarrow\quad
    \arg\max_i z_i^{(b)}=\arg\max_i z_i^{\star}.
    \label{eq:top1-certificate}
\end{equation}
A large fresh margin can therefore tolerate more residual cache influence,
whereas a low-margin decision may require a longer recovery window even under
the same adapter update.  This helps explain why recovery behavior depends not
only on the updated module, layer, magnitude, and version gap, but also on the
input and the model's current decision geometry.

Together, these conditions characterize when direct reuse or bounded recovery
is sufficient, and when complete affected-suffix recomputation remains
necessary.  Exact $T_b$ requires counterfactual
hybrid executions and is therefore an offline analysis quantity rather than a
deployable runtime certificate.  The next section introduces the empirical
runtime policy derived from this analysis.  Detailed dependency derivations
and proofs appear in the supplementary material.


\section{Methodology}

\begin{figure*}[t]
    \centering
    \includegraphics[width=0.95\textwidth]{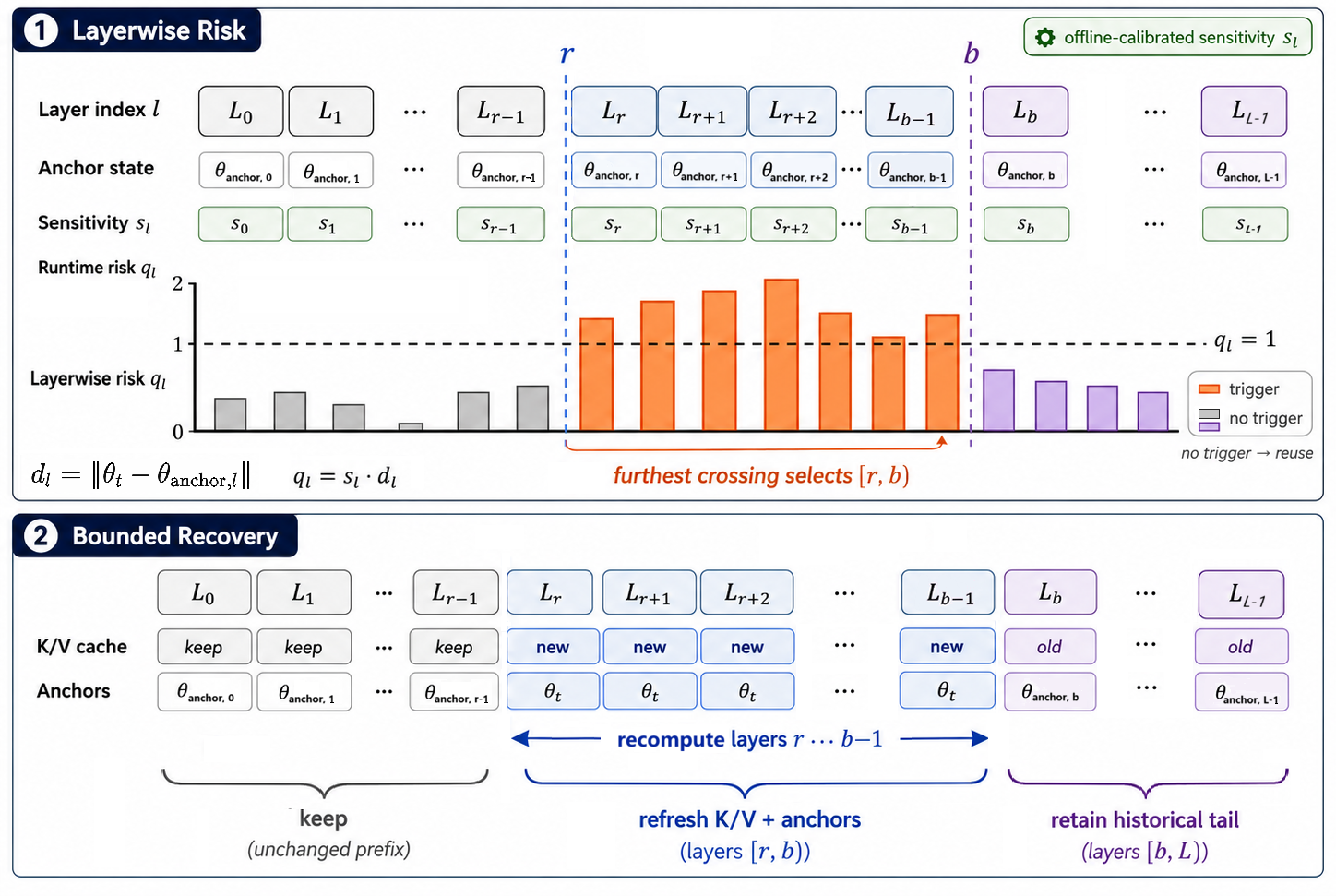}
    \caption{Overview of \textsc{CacheReforge} workflow.}
    \label{fig:cachereforge-workflow}
\end{figure*}

\textsc{CacheReforge} recovers stale KV caches as the same adapter evolves.
Given an adapter update, it locates the first executable restart layer,
estimates layerwise cache risk, and selects a contiguous recovery range.  Each
cached layer stores a per-layer adapter anchor, making its validity depend on
the model state that produced it rather than on token and context identity alone.
Independently updating these anchors tracks the mixed-version cache created by
repeated partial recovery.  Offline-calibrated layer sensitivities then combine
with runtime adapter drift to distinguish direct reuse, bounded recovery, and
full affected-suffix recomputation.  The selected range is executed from the
restart layer using retained boundary states, after which its K/V states and
adapter anchors are refreshed.  Figure~\ref{fig:cachereforge-workflow}
summarizes this workflow.

\subsection{Layerwise Risk}

\noindent\textbf{Offline calibration.} For each fixed deployment configuration, a disjoint calibration split provides trajectories. For condition $i$ with restart layer $r_i$, we evaluate
$\mathcal H_i=\{\min(h,N-r_i):h\in\{0,1,2,4,8,16\}\}\cup\{N-r_i\}$.
Here, $h=0$ denotes direct reuse, positive $h$ recomputes that many contiguous layers from $r_i$, and $N-r_i$ is the complete affected suffix. Relative to fresh full-prefill recomputation under the current adapter, a candidate is safe when first-step next-token KL is at most $0.05$, top-1 agreement is at least $0.99$, and full-answer task-score drop is at most $0.01$. The first safe window, $h_i^{\star}=\min\{h\in\mathcal H_i:S_i(h)=1\}$, is an empirical calibration target; larger evaluated windows need not remain safe.

The selected window induces the per-layer binary calibration label
\begin{equation}
    y_{i,\ell}=\mathbb{I}[r_i\leq \ell<r_i+h_i^{\star}],
    \label{eq:layer-recovery-label}
\end{equation}
which indicates whether layer $\ell$ lies in the empirically selected range under the observed adapter displacement. Let $d_i$ be the measured $\ell_2$ distance between the active adapter and the adapter state associated with the stale calibration cache. For each layer, form $\mathcal D_{\ell}^{+}=\{d_i:y_{i,\ell}=1\}$ of size $n_\ell$, and let $d_{\ell,(k)}$ denote its $k$th order statistic. For $n_\ell>0$, the frozen code sets $\delta_\ell=\max\{d_{\ell,(\lfloor0.05n_\ell\rfloor+1)},10^{-12}\}$, the first value after the allowed $5\%$ lower tail. The sensitivity is
\begin{equation}
    s_\ell=
    \begin{cases}
        1/\delta_\ell, & \mathcal D_{\ell}^{+}\neq\emptyset,\\
        0, & \mathcal D_{\ell}^{+}=\emptyset.
    \end{cases}
    \label{eq:layer-sensitivity}
\end{equation}
Thus, layers included in selected ranges after small parameter changes receive larger sensitivities. Indexed by model, update target, adapter configuration, context regime, and workload, the resulting profile is frozen before held-out evaluation.

\noindent\textbf{Runtime risk and range selection.} A cache may mix layers refreshed at different adapter versions, so \textsc{CacheReforge} maintains a separate adapter anchor $\theta_{\ell}^{\mathrm{anchor}}$ for each cached layer. After the adapter changes to $\theta_t$, layer $\ell$ has displacement and calibrated risk
\begin{equation}
    d_\ell(t)=
    \left\|\theta_t-\theta_{\ell}^{\mathrm{anchor}}\right\|_2,
    \qquad
    q_\ell(t)=s_\ell d_\ell(t).
    \label{eq:layer-runtime-risk}
\end{equation}
A layer is triggered when $q_\ell(t)\geq 1$. Unlike version age, this risk uses the actual parameter displacement from the state that produced the layer's current cache; consecutive updates can therefore accumulate risk at different rates, while partial cancellation can reduce it.

If no layer is triggered, \textsc{CacheReforge} directly reuses the cache; otherwise, it chooses
\begin{equation}
    b_t=1+\max\{\ell:q_\ell(t)\geq 1\}
    \label{eq:recovery-window-end}
\end{equation}
and recomputes the minimal executable contiguous range $[r,b_t)$. For an update at layer $s$, key or value updates can directly invalidate layer $s$; query, output, and MLP updates first affect the cache at $s+1$, although recovery still restarts from $s$ whenever downstream states are affected. The furthest layer whose calibrated drift risk crosses the threshold sets the endpoint: $b_t=N$ is an exact affected-suffix restart; otherwise, recovery is bounded.

\subsection{Bounded Recovery}

\noindent\textbf{Executable restart.} During the original prefill, \textsc{CacheReforge} retains the hidden state at supported
restart boundaries in addition to the KV cache.  To execute a recovery window,
it supplies the stored boundary state before layer $r$ to the decoder under the
current adapter, processes the complete prefix through layers $r$ to $b_t-1$,
and overwrites the K/V entries produced by those layers.  Layers before $r$ are
unaffected by the update and remain unchanged; layers from $b_t$ onward retain
their historical cache states.  The normalized recomputation fraction is
therefore $(b_t-r)/N$.

\noindent\textbf{Cache and anchor updates.} After refreshing $[r,b_t)$, the system sets
$\theta_{\ell}^{\mathrm{anchor}}\leftarrow\theta_t$ only for layers inside
that range.  Each layer's anchor changes only when that layer is recomputed, so
the runtime risk vector evolves with the mixed-version cache itself.  Unrecomputed
layers retain their older anchors and continue to accumulate displacement
relative to the adapter state that generated their cached K/V tensors.

The cache preserves per-layer version and anchor metadata, so bounded recovery
is not relabeled as fully current while its unrecomputed tail remains historical.
All refreshed states are produced by ordinary decoder execution under the active
adapter; the method performs no post-hoc tensor correction or access to
fresh-reference states.  This makes the runtime decision inexpensive while
retaining the layerwise structure revealed by the recomputation-horizon analysis.

\section{Experiments}

We conduct this study by investigating the following research questions.  \textbf{RQ1: Does adapter evolution make stale KV caches unsafe?}
We first establish whether previously computed KV states become meaningfully
inconsistent with the current model.  \textbf{RQ2: What recovery outcomes arise,
and how much recomputation do they require?}  We then measure the empirical
recomputation horizon across update conditions, distinguishing safe direct
reuse, bounded recomputation, and full affected-suffix recomputation.  \textbf{RQ3: What determines the recovery
boundary?}  Because the required range varies across updates, we test whether
cumulative unresolved-suffix proxies motivated by tail influence predict this
variation.  \textbf{RQ4: Does
drift-sensitive recovery improve the quality--cost trade-off?}  Finally, we use
the resulting signal to guide recovery over consecutive updates and evaluate
its end-to-end practical value.

\subsection{Experimental Setup}

\paragraph{Models.}
We use Qwen2.5-1.5B-Instruct for dense mechanism analysis and
Qwen2.5-7B-Instruct for realistic validation \citep{qwen2024qwen25}.
Qwen2.5 is a widely used decoder-only Transformer with standard KV caching,
making it a practical testbed for controlled stale-cache analysis.

\paragraph{Datasets.}
Our controlled studies use 4K multi-hop tracing and 16K variable-tracking
workloads.  Realistic evaluation uses the HotpotQA \citep{yang2018hotpotqa} and
2WikiMQA \citep{ho2020constructing} subsets of LongBench
\citep{bai2024longbench} at 16K context.

\paragraph{Implementation.}
We freeze the base model and evolve a rank-8 LoRA adapter through
answer-supervised updates, using bfloat16 arithmetic throughout.  Fresh
full-prefill recomputation under the active adapter defines reference behavior.
We provide the implementation, experiment configurations, and complete result
artifacts in the accompanying reproducibility package.

\subsection{RQ1: Adapter Evolution Makes Stale KV Caches Unsafe}

\textbf{Design.}
To quantify whether adapter evolution makes cached states unsafe, we compare
stale reuse with fresh recomputation on Qwen2.5-7B at version gaps
$\{0,1,2,4,8,16,32\}$ over three paired seeds.  Controlled runs use
normalized update magnitude $10^{-3}$, vary the updated projection and layer
position, and contain 64 samples per seed.  HotpotQA and 2WikiMQA use
value-projection updates at magnitude $10^{-2}$ and 32 samples per seed.  We
measure first-step next-token KL and full-answer task score.

\begin{figure}[t]
    \centering
    \includegraphics[width=\columnwidth]{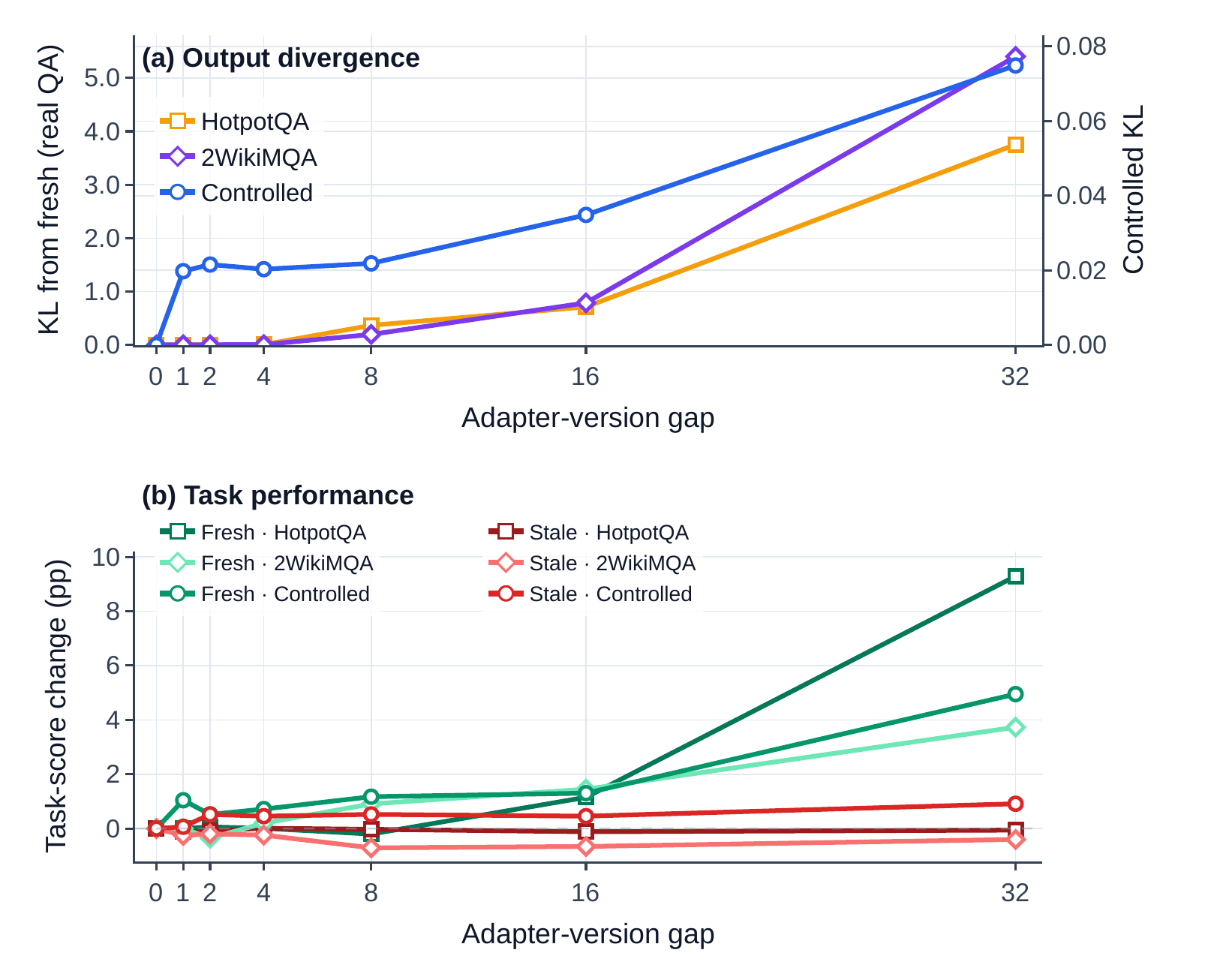}
    \caption{Adapter evolution and stale-KV mismatch on Qwen2.5-7B.  Top: KL
    divergence between stale reuse and fresh recomputation.  Bottom: task-score
    change from each workload's pre-update baseline.}
    \label{fig:stale-drift}
\end{figure}

\textbf{Results.} Figure~\ref{fig:stale-drift} compares stale reuse with fresh recomputation.  On the controlled workload, mean KL rises from $0.0198$ after one update to $0.0749$ at version gap 32.  The lower panel reports task-score changes from each pre-update baseline.  At gap 32, fresh recomputation improves the controlled score from $53.84\%$ to $58.79\%$, whereas stale reuse reaches only $54.75\%$, erasing $4.04$ of the $4.95$-point adaptation gain.  Despite small-gap fluctuations, the long-gap separation is clear.

Sensitivity also depends strongly on the updated module: averaged over all
tested gaps, value and joint query--value updates produce the largest KL
divergences ($0.0754$ and $0.0672$) and task-score losses ($3.39$ and $2.86$
points), respectively.  Updates near or outside the cache-sensitive path can
have negligible cache-induced error, confirming that version mismatch alone
does not determine severity.

The realistic QA tasks exhibit the same failure more sharply.  KL remains below
$0.007$ through gap 4 but rises to $3.753$ on HotpotQA and $5.401$ on 2WikiMQA
at gap 32.  At that gap, fresh recomputation improves token-F1 from $7.07\%$ to
$16.36\%$ and from $9.07\%$ to $12.80\%$, whereas stale reuse yields only
$7.01\%$ and $8.67\%$, respectively.

\finding{Adapter evolution can invalidate historical KV states, and direct
stale reuse can erase most adaptation gains, especially as the version gap
grows.}

\subsection{RQ2: Recovery Has Three Distinct Outcomes}

\textbf{Design.}
We measure empirical recomputation horizons on Qwen2.5-1.5B across eight
target--position combinations, version gaps $\{1,4,16\}$, three seeds, and
windows of 1, 2, 4, 8, 16, and the complete affected suffix, yielding 72
profiles of 16 paired samples.  After testing $h=0$, we select the smallest
positive executable window with aggregate KL $\leq0.05$, top-1 agreement
$\geq0.99$, task-score drop $\leq0.01$, and no paired sample violating any
threshold.  This first-hit statistic does not assume monotonic safety.  Complete suffixes contain
28, 14, and 1 layer for early, middle, and late updates, respectively.

\begin{figure}[t]
    \centering
    \includegraphics[width=\columnwidth]{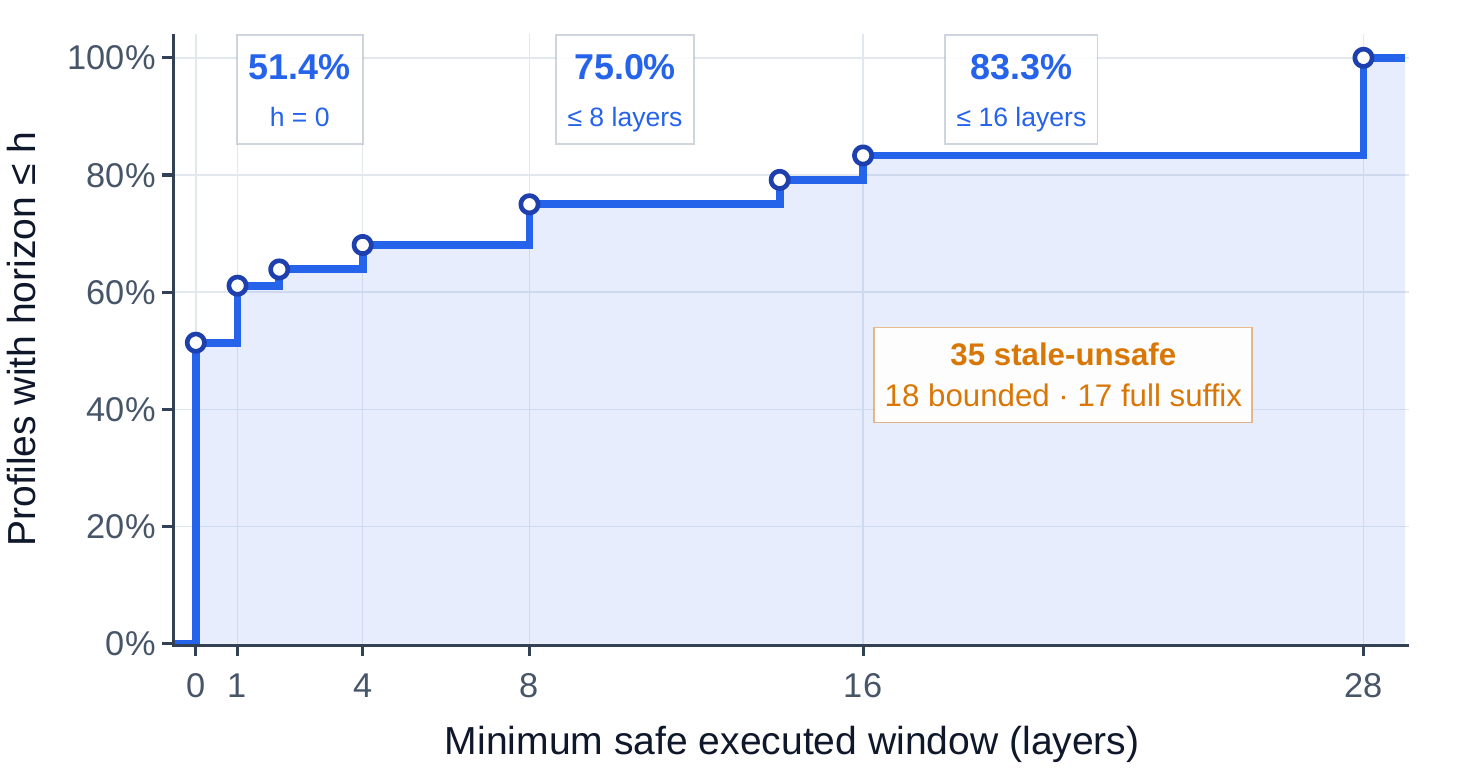}
    \caption{Empirical recomputation horizons.}
    \label{fig:horizon-cdf}
\end{figure}

\textbf{Results.}
Figure~\ref{fig:horizon-cdf} shows three outcomes.  Direct reuse is safe for 37
of 72 profiles ($51.4\%$).  Among the 35 stale-unsafe profiles, 18 ($51.4\%$)
recover before the affected-suffix end and 17 ($48.6\%$) require the full
suffix: twelve 28-layer early, three 14-layer middle, and two one-layer late-key
profiles.  Positive bounded horizons span 1--16 executed layers.

These results also confirm the module-dependent invalidation onset detailed in
the supplement.  For query and MLP updates, one-layer execution repairs no
affected KV state and exactly matches stale reuse across all 288 query and 432
MLP rows; positive one-layer recovery occurs only for key or value updates.
Outcomes remain condition-dependent: six of nine early-key profiles admit
bounded positive recovery, five of nine early-MLP profiles require the full
suffix, and middle-query profiles split into one no-recovery, six bounded, and
two full-suffix cases.  Full-suffix recovery occurs in eight of 24 profiles at
gap 16, versus five at gap 1 and four at gap 4.  Moreover, 50 of 72 profiles
contain a local KL reversal as the window grows.

\finding{Recovery has three condition-dependent empirical outcomes: direct
reuse, a bounded first acceptable window, or the full affected suffix; larger
windows do not guarantee improvement.}

\subsection{RQ3: Tail-Influence Proxies Predict Recovery}

\textbf{Design.}
To identify what determines the recovery boundary, we compare window KL with
layerwise drift, cumulative stale-suffix error, boundary error, and final
suffix-attention input error across the 72 RQ2 profiles.

\begin{figure}[t]
    \centering
    \includegraphics[width=\columnwidth]{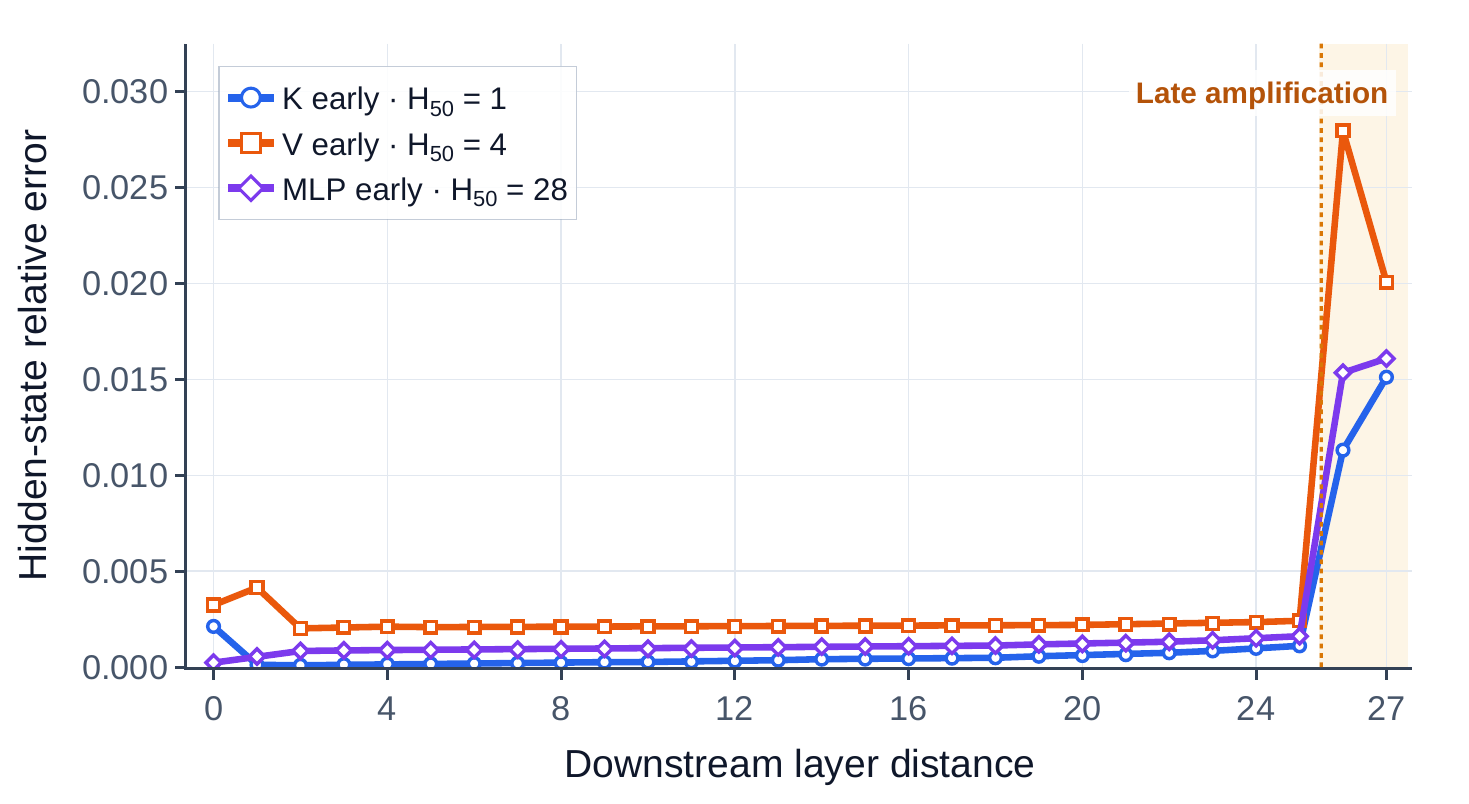}
    \caption{Hidden-state propagation after early-layer updates.}
    \label{fig:propagation}
\end{figure}

\textbf{Results.}
The propagation measurements do not support a universal attenuation law.
Across the 72 frozen profiles, hidden-state drift often remains small through
much of the network and reaches its maximum near the final layers; the automatic
profile analysis classifies every condition as late amplification.
Figure~\ref{fig:propagation} shows mean trajectories for three early-layer
update targets, averaged over three seeds and three version gaps; $H_{50}$
denotes the median safe horizon.  Early value updates produce the largest peak
drift, whereas early key updates remain lower despite sharing the same
dependency depth.

Propagation proxies motivated by the tail-influence view are more predictive
than a single boundary measurement.  For every positive window, we join its
remaining stale suffix with the W2 layerwise propagation trace.  The sum of
hidden-state errors over that suffix has Spearman correlation $0.873$ with
window KL overall and $0.690$ after rank-centering within each
seed--target--gap condition.  The corresponding single boundary error reaches
only $0.666$ overall and $0.320$ within condition.  Key- and value-tail sums
show the same within-condition correlation of $0.690$.

W3 provides an independent check closer to the output path.  The final
unrepaired suffix-attention input error, an internal-state proxy rather than the
output-space $T_b$ in Equation~\ref{eq:tail-influence}, correlates with KL at
$0.892$ overall and $0.620$ within condition.  A log--log model with seed,
update target, and version gap removed as fixed effects gives
\begin{equation}
    D_{\mathrm{KL}} \propto E_{\mathrm{tail}}^{2.14},
    \qquad R^2_{\mathrm{within}}=0.115.
    \label{eq:empirical-quadratic-tail}
\end{equation}
The near-quadratic exponent is qualitatively aligned with the bound's quadratic
dependence, but it does not directly verify
Equation~\ref{eq:tail-kl-certificate} because $E_{\mathrm{tail}}$ is a proxy
rather than $T_b$.  The offline proxy is not a runtime certificate.

\finding{Cumulative unresolved-suffix influence proxies predict recovery better
than a boundary measurement, motivating a layerwise risk proxy.}

\begin{table*}[t]
\centering
\small
\setlength{\tabcolsep}{5.0pt}
\begin{tabular}{l|rrr|rr}
\toprule
Method & HotpotQA (\%) $\uparrow$ & 2WikiMQA (\%) $\uparrow$ & KL $\downarrow$ & Recomp. (\%) $\downarrow$ & Maint. (s) $\downarrow$ \\
\midrule
Fresh full & 9.41 & 9.40 & 0.0000 & 100.00 & 50.26 \\
Exact affected suffix & 9.41 & 9.40 & 0.0000 & 50.00 & 28.75 \\
\textsc{CacheReforge} & 9.13 & 9.16 & 0.0335 & 5.44 & 3.42 \\
Stale reuse & 7.95 & 9.03 & 0.4385 & 0.00 & 0.31 \\
\bottomrule
\end{tabular}
\caption{Held-out 32-version quality--maintenance-cost comparison on HotpotQA and
2WikiMQA.}
\label{tab:end-to-end}
\end{table*}

\subsection{RQ4: Drift-Sensitive Recovery Improves the Quality--Cost Trade-off}

\textbf{Design.}
To assess the practical quality--cost trade-off, we evaluate recovery over 32
consecutive middle value-projection updates of magnitude $10^{-2}$ on
Qwen2.5-7B.  Two calibration seeds use eight samples per task, and three disjoint
evaluation seeds use 16, totaling 96 trajectories and 3,072 cache decisions per
method.  We compare fresh full, exact affected suffix, stale reuse,
\textsc{CacheReforge}, and two policy ablations on HotpotQA and 2WikiMQA.
\emph{Drift-only} removes layerwise sensitivity and applies one
calibration-frozen drift threshold to all affected layers.  \emph{Fixed-window}
refreshes the same two-layer window after every update, giving a slightly larger
average recomputation budget than \textsc{CacheReforge}.  All methods observe the same
updates and held-out requests.  \textsc{CacheReforge} updates only the layer anchors
covered by each recovery action; no policy accesses fresh logits or hidden
states at runtime.

\textbf{Results.} Table~\ref{tab:end-to-end} shows that exact affected-suffix
recomputation matches fresh behavior at $50.00\%$ recomputation.
\textsc{CacheReforge} uses $89.1\%$ less recomputation than this exact baseline,
reduces mean KL by $92.4\%$ relative to stale reuse, raises top-1 agreement from
$84.47\%$ to $94.04\%$ at $5.44\%$ recomputation, and cuts cache-maintenance
time by $93.2\%$ relative to full prefill.  Within the shared held-out runs,
these prefill-side savings reduce mean total request time from $4.16$~s per
update for fresh full recomputation to $2.30$~s for \textsc{CacheReforge}, a
$44.7\%$ reduction.

\begin{table}[t]
\centering
\small
\setlength{\tabcolsep}{4.0pt}
\begin{tabular}{lrrr}
\toprule
Method & KL $\downarrow$ & Top-1 (\%) $\uparrow$ & Recomp. (\%) $\downarrow$ \\
\midrule
\textsc{CacheReforge} & 0.0335 & 94.04 & 5.44 \\
Drift-only & 0.0581 & 92.22 & 3.82 \\
Fixed-window & 0.2198 & 88.05 & 7.14 \\
\bottomrule
\end{tabular}
\caption{Recovery-policy ablations on the frozen RQ4 trajectories.}
\label{tab:rq4-ablation}
\end{table}

Table~\ref{tab:rq4-ablation} isolates the two recovery-policy components.  Removing
layerwise sensitivity reduces fidelity, whereas the fixed window is less
faithful despite its larger budget.  Task-specific recomputation is $6.25\%$ on
HotpotQA and $4.63\%$ on 2WikiMQA.

\finding{\textsc{CacheReforge} remains approximate, but approaches fresh quality with
$5.44\%$ mean recomputation; both layerwise sensitivity and adaptive range
selection improve fidelity.}

\section{Related Work}

\paragraph{KV cache management and reuse.}
Existing systems reduce KV cache costs through efficient allocation, reuse,
movement, and compression.  PagedAttention virtualizes cache storage; Prompt
Cache and SGLang reuse prefix states across requests; LMCache extends reuse
across engines and storage tiers; and KIVI applies low-bit quantization
\citep{kwon2023vllm,gim2024promptcache,zheng2024sglang,cheng2025lmcache,liu2024kivi}.
CacheBlend selectively refreshes cached states when separately processed context
chunks are composed \citep{yao2025cacheblend}.  These methods generally assume
fixed model parameters and do not address invalidation caused by adapter
evolution.

\paragraph{Adapter-aware serving.}
Punica and SLoRA batch requests across LoRA variants and manage adapter and cache
memory \citep{chen2024punica,sheng2024slora}.  Activated LoRA preserves a
base-model prefix cache by restricting adaptation to later tokens, while LRAgent
and ForkKV share or copy KV states across distinct LoRA agents
\citep{greenewald2025alora,jeon2026lragent,wang2026forkkv}.  These approaches
serve stable adapter identities or preserve a stable shared cache component,
rather than recover caches after repeated updates to the same adapter.

\paragraph{Test-time adaptation and online learning.}
Dynamic evaluation, test-time training, Tent, and CoTTA adapt model parameters
from recent or unlabeled test inputs
\citep{krause2018dynamic,sun2020ttt,wang2021tent,wang2022cotta}.  Recent LLM
methods use lightweight LoRA updates or problem-specific reinforcement learning
during generation \citep{hu2025tlm,yuksekgonul2026tttdiscover}.  Most closely
related, qTTT updates only query projections after a single prefill so that the
existing KV cache remains valid \citep{bansal2026qttt}; it therefore avoids the
cache-recovery problem created by updates to cache-producing modules.

\paragraph{Novelty of Our Study.}
Overall, prior work either reuses caches under fixed parameters, serves
multiple stable adapters, or adapts models without recovering KV states affected
by the update.  Our study connects these previously separate concerns by
treating cache validity as jointly determined by the token prefix and the
evolving adapter version.  It introduces a layerwise mixed-version cache
representation, distinguishes global dependency depth from the functional
recomputation horizon, and characterizes recovery through cumulative tail
influence.  Building on this formulation, \textsc{CacheReforge} performs
drift-sensitive bounded recomputation with an exact-suffix fallback, while our
controlled and realistic evaluations establish when recovery is necessary,
when it can remain bounded, and how much computation it saves relative to fresh
execution.

\section{Conclusion}

We study stale KV cache recovery under evolving parameter efficient adapters. \textsc{CacheReforge} combines module aware restart boundaries, calibrated sensitivity, and adapter drift to select direct reuse, bounded recomputation, or complete affected suffix recovery. Controlled and long context evaluations show that it reduces output divergence while preserving most cached computation. Its effectiveness depends on model architecture, task, update rule, tolerance, generation regime, and restart state support.\looseness=-1

\bibliographystyle{plainnat}
\bibliography{references}

\par\bigskip
\setcounter{secnumdepth}{2}
\setcounter{section}{0}
\setcounter{subsection}{0}
\setcounter{equation}{0}
\setcounter{table}{0}
\setcounter{figure}{0}
\renewcommand{\thesection}{S\arabic{section}}
\renewcommand{\thesubsection}{\thesection.\arabic{subsection}}
\renewcommand{\theequation}{S\arabic{equation}}
\renewcommand{\thetable}{S\arabic{table}}
\renewcommand{\thefigure}{S\arabic{figure}}

\begin{center}
{\Large\bfseries Supplementary Material}
\end{center}

\begin{table*}[t]
\centering
\small
\setlength{\tabcolsep}{5pt}
\renewcommand{\arraystretch}{1.08}
\begin{tabular}{@{}p{0.22\textwidth}p{0.73\textwidth}@{}}
\toprule
Term or symbol & Meaning \\
\midrule
Adapter evolution & Successive updates to the same lightweight adapter while the base model remains frozen. \\
Stale KV cache & $\mathcal C^{(t)}$ is generated under $v_t$ but retained under $v_{t+\Delta}$; the prefix is unchanged, but the cache and current model state are mismatched. \\
Recovery modes & Direct reuse executes $h=0$ layers; bounded recovery has $0<h<N-r(c)$; exact affected-suffix recomputation executes the complete executable suffix, $h=N-r(c)$; fresh full recomputation reruns all $N$ layers and defines the reference. \\
Dependency depth vs. horizon & Dependency depth is the exact downstream propagation extent. $H_{\epsilon}(c)$ is the minimum executable recovery length meeting population tolerance $\epsilon$; the RQ2 empirical horizon is the first tested candidate passing both aggregate and per-sample criteria. \\
$N,\ell,s,\Delta$ & Number of layers, generic layer index, updated layer, and adapter version gap. \\
$r_{\mathrm{KV}}(m,s),r(c)$ & Earliest cached layer that can change after updating module $m$ at layer $s$, and earliest executable restart layer. They can differ for query, output, and MLP updates. \\
$h,b$ & Executed recovery length and exclusive layer boundary. In the hybrid analysis, layers $[b,N)$ remain historical. \\
$p_c^{\star},p_{c,h}$ & Fresh-reference distribution and distribution after executing $h$ recovery layers for condition $c$. \\
$z^{\star},z^{(b)}$ & Fresh-reference logits and hybrid logits with refreshed states before $b$ and historical states from $b$ onward. \\
$v_{\ell},T_b$ & Incremental logit influence of refreshing layer $\ell$ and cumulative influence magnitude of the unresolved suffix beginning at $b$. \\
$\theta_t,\theta_{\ell}^{\mathrm{anchor}}$ & Current adapter state and the adapter state that produced the cache currently stored at layer $\ell$. \\
$s_{\ell},d_{\ell}(t),q_{\ell}(t)$ & Calibrated sensitivity, displacement from the layer anchor, and runtime risk $q_{\ell}(t)=s_{\ell}d_{\ell}(t)$. \\
$\mathcal T_t,b_t$ & Triggered layer set and, when nonempty, exclusive runtime recovery end $b_t=1+\max\mathcal T_t$. \\
\bottomrule
\end{tabular}
\caption{Reference terminology and notation. All layer ranges are half-open.}
\label{tab:sup-terminology}
\end{table*}

\section{Detailed Recovery Analysis}

This section gives the formal recovery definition and the proofs underlying
the theoretical analysis in the main paper.  To avoid ambiguity between a layer index and the
number of layers, let the Transformer contain $N$ layers indexed
$0,\ldots,N-1$ throughout this section.

\subsection{Invalidation and Functional Recovery}

\paragraph{Module-dependent onset.}

Let $H_{\ell}^{(v)}$ denote the prefix hidden states after Transformer layer
$\ell$ under adapter version $v$, and let
$C_{\ell}^{(v)}=(K_{\ell}^{(v)},V_{\ell}^{(v)})$ be the cache written by that
layer.  We use $H_{-1}^{(v)}$ for the embedding/input state before layer $0$.
Abstractly,
\begin{equation}
    H_{\ell}^{(v)} = F_{\ell}\!\left(H_{\ell-1}^{(v)};
    \theta_{\ell}^{(v)}\right), \qquad
    C_{\ell}^{(v)} = G_{\ell}\!\left(H_{\ell-1}^{(v)};
    \theta_{\ell}^{(v)}\right).
    \label{eq:sup-layer-cache-dependency}
\end{equation}
If an update changes layer $s$, it may perturb $H_s$ and every later layer.  A
first-order expansion gives
\begin{equation}
    \delta H_{\ell} \approx
    J_{\ell}J_{\ell-1}\cdots J_{s+1}\delta H_s,
    \qquad \ell>s,
    \label{eq:sup-downstream-propagation}
\end{equation}
where $J_i$ is the local Jacobian of layer $i$.  Since this product is generally
nonzero, exact consistency conservatively requires executing the complete
downstream dependency closure.

For a standard decoder block, let $\bar H_{s-1}$ denote the normalized input to
self-attention.  Suppressing positional rotation and bias terms, which do not
affect the dependency argument,
\begin{equation}
    \begin{aligned}
        Q_s&=\bar H_{s-1}W_s^Q,\\
        K_s&=\bar H_{s-1}W_s^K, \qquad
        V_s=\bar H_{s-1}W_s^V.
    \end{aligned}
    \label{eq:sup-qkv-current-layer}
\end{equation}
A key- or value-projection update can directly change the cache written at layer
$s$.  A query, attention-output, or MLP update changes the layer output while
leaving the keys and values already computed at that same layer unchanged.
Hence the earliest cache layer that can change is
\begin{equation}
    r_{\mathrm{KV}}(m,s)=
    \begin{cases}
        s, & m\in\{K,V\},\\
        \min\{s+1,N\}, & m\in\{Q,O,\mathrm{MLP}\}.
    \end{cases}
    \label{eq:sup-first-invalid-cache-layer}
\end{equation}
Here $N$ is an end-boundary sentinel indicating that no cached layer is
invalidated.  This occurs, for example, after a query, output, or MLP update in
the final Transformer layer.
For query, output, and MLP updates, an executable restart must nevertheless
begin at layer $s$ whenever a later cache layer can be affected.  Executing only
that layer replaces no affected KV state, so a one-layer operation is equivalent
to stale reuse with respect to the cache subsequently consumed during decoding.

\paragraph{Functional recovery and recomputation horizon.}

Let $c$ denote an update condition and $r(c)$ its earliest executable restart
layer.  Define $\mathcal{R}_{r,h}$ as an operation that executes the half-open
range $[r,r+h)$ under the current adapter, writes the resulting KV states, and
retains historical states outside that range.  The case $h=0$ is direct stale
reuse; $h=N-r$ executes the complete executable suffix.  For input $x$ at a
fixed decoding step, let $p_{c,h}(\cdot\mid x)$ be the resulting next-token
distribution and $p_c^{\star}(\cdot\mid x)$ the fresh-recomputation reference.
For discrepancy
$\mathcal D$, define
\begin{equation}
    \mathcal{E}_c(h)=
    \mathbb{E}_{x\sim\mathcal{X}_c}
    \left[\mathcal{D}\!\left(p_c^{\star}(\cdot\mid x),
    p_{c,h}(\cdot\mid x)\right)\right].
    \label{eq:sup-functional-recovery-error}
\end{equation}
The recomputation horizon at tolerance $\epsilon$ is
\begin{equation}
    H_{\epsilon}(c)=
    \min\left\{h\in\{0,\ldots,N-r(c)\}:\
    \mathcal{E}_c(h)\leq\epsilon\right\}.
    \label{eq:sup-recomputation-horizon}
\end{equation}
A condition may therefore require no recovery, bounded positive recovery, or
the complete executable suffix.

Equation~\ref{eq:sup-recomputation-horizon} defines a population discrepancy
horizon.  The RQ2 tables report an empirical operational horizon over the tested
candidate windows: the smallest candidate whose aggregate KL, top-1 agreement,
and task-score drop satisfy the thresholds and for which no paired sample
violates any threshold.  This is a first-hit statistic and does not imply that
the finite-sample criterion is monotone across larger candidates.  The two
quantities express the same recovery concept but use different safety predicates
and should not be identified numerically.

\subsection{Tail-Influence Certificates}

\paragraph{Decomposition.}

Fix an input $x$ and omit it from the notation below.  Let $z^{\star}$ be the
fresh next-token logits at a fixed decoding step; the
reported measurements use the first generated-token step.  For the offline
hybrid sequence, let $z^{(b)}$ use refreshed cache states before boundary $b$
and historical states from $b$ through layer $N-1$, with
$z^{(N)}=z^{\star}$.  Define
\begin{equation}
    v_{\ell}=z^{(\ell)}-z^{(\ell+1)},
    \qquad
    T_b=\sum_{\ell=b}^{N-1}\lVert v_{\ell}\rVert_2.
    \label{eq:sup-tail-influence}
\end{equation}
The difference from fresh recomputation telescopes exactly:
\begin{align}
    z^{(b)}-z^{\star}
    &=z^{(b)}-z^{(N)} \\
    &=\sum_{\ell=b}^{N-1}
      \left(z^{(\ell)}-z^{(\ell+1)}\right)
      =\sum_{\ell=b}^{N-1}v_{\ell}.
    \label{eq:sup-tail-decomposition}
\end{align}
The triangle inequality then gives
\begin{equation}
    \lVert z^{(b)}-z^{\star}\rVert_2
    \leq\sum_{\ell=b}^{N-1}\lVert v_{\ell}\rVert_2=T_b.
    \label{eq:sup-tail-norm-bound}
\end{equation}
This bound does not assume layerwise contraction.  It remains valid when
individual influences amplify or cancel.

\paragraph{KL certificate.}

Let $A(z)=\log\sum_i e^{z_i}$.  Its gradient is the softmax, and
\begin{equation}
    \nabla^2 A(z)=\operatorname{diag}(p)-pp^{\top},
    \qquad p=\operatorname{softmax}(z).
    \label{eq:sup-lse-hessian}
\end{equation}
For any unit vector $u$, the quadratic form
$u^{\top}(\operatorname{diag}(p)-pp^{\top})u$ is the variance of a categorical
random variable taking value $u_i$.  Popoviciu's inequality bounds this variance
by one quarter of the squared range, and
$\max_i u_i-\min_i u_i\leq\sqrt{2}\lVert u\rVert_2=\sqrt{2}$.
Therefore $\lVert\nabla^2 A(z)\rVert_2\leq 1/2$, so $A$ is $1/2$-smooth in the
Euclidean norm.

Write $\Delta_b=z^{(b)}-z^{\star}$.  The softmax KL divergence is the
Bregman divergence of $A$:
\begin{align}
    D_{\mathrm{KL}}\!\left(p^{\star}\Vert p^{(b)}\right)
    &=A(z^{(b)})-A(z^{\star})
      -\langle\nabla A(z^{\star}),\Delta_b\rangle \\
    &\leq \frac{1}{4}\lVert\Delta_b\rVert_2^2 \\
    &\leq \frac{1}{4}T_b^2.
    \label{eq:sup-tail-kl-certificate}
\end{align}
The first inequality follows from $1/2$-smoothness, and the second from
Equation~\ref{eq:sup-tail-norm-bound}.

Restoring the input dependence, write $T_b(x)$ for the corresponding tail
influence.  When $\mathcal D$ in
Equation~\ref{eq:sup-functional-recovery-error} is the forward KL used in the
experiments, the pointwise certificate implies
\begin{equation}
    \mathcal E_c(h)
    \leq \frac{1}{4}\,
    \mathbb E_{x\sim\mathcal X_c}
    \left[T_{r(c)+h}(x)^2\right].
    \label{eq:sup-expected-tail-kl-certificate}
\end{equation}
Thus $\mathbb E_x[T_{r(c)+h}(x)^2]\leq4\epsilon$ is sufficient for the
population KL horizon in Equation~\ref{eq:sup-recomputation-horizon}; the
stronger pointwise condition $T_{r(c)+h}(x)\leq2\sqrt{\epsilon}$ gives the same
guarantee for every covered input.

\paragraph{Top-1 certificate.}

Let $y^{\star}=\arg\max_i z_i^{\star}$ and define the fresh margin
\begin{equation}
    \gamma^{\star}=z_{y^{\star}}^{\star}
    -\max_{j\neq y^{\star}}z_j^{\star}.
    \label{eq:sup-fresh-margin}
\end{equation}
Write $\delta=z^{(b)}-z^{\star}$.  For any $j\neq y^{\star}$,
\begin{align}
    z_{y^{\star}}^{(b)}-z_j^{(b)}
    &=(z_{y^{\star}}^{\star}-z_j^{\star})
      +(\delta_{y^{\star}}-\delta_j) \\
    &\geq \gamma^{\star}-2\lVert\delta\rVert_{\infty}
     \geq \gamma^{\star}-2\lVert\delta\rVert_2
     \geq \gamma^{\star}-2T_b.
    \label{eq:sup-top1-proof}
\end{align}
Consequently, $T_b<\gamma^{\star}/2$ makes every pairwise margin positive and
preserves the fresh top-1 prediction.  The condition is sufficient rather than
necessary and is intentionally conservative.

\subsection{Non-Monotonicity and Condition Dependence}

Although $T_b$ decreases as terms are removed, the actual logit error need not
be monotonic.  Let
$R_b=z^{(b)}-z^{\star}=v_b+R_{b+1}$.  Refreshing layer $b$ changes the squared
error by
\begin{equation}
    \lVert R_{b+1}\rVert_2^2-\lVert R_b\rVert_2^2
    =-\lVert v_b\rVert_2^2-2\langle v_b,R_{b+1}\rangle.
    \label{eq:sup-cache-cancellation}
\end{equation}
The error can temporarily increase when $v_b$ previously cancelled part of the
residual influence of later stale layers.

An update condition can be summarized as
\begin{equation}
    c=(m,s,\Delta,\rho,\xi),
    \label{eq:sup-update-condition}
\end{equation}
where $m$ is the updated module, $s$ its layer position, $\Delta$ the adapter
version gap, $\rho$ the update magnitude, and $\xi$ collects model- and
workload-level factors, such as the distributions of attention patterns and
output margins under $\mathcal X_c$.  These variables alter both the injected
perturbation and the alignment among the influence vectors.  A single
global recomputation window can therefore over-compute low-risk conditions while
under-repairing persistent ones.

\section{Additional Experimental Details}

\subsection{Protocol}

RQ1 uses Qwen2.5-7B-Instruct with version gaps
$\{0,1,2,4,8,16,32\}$ and three paired seeds.  The controlled workload uses
64 samples per seed; HotpotQA and 2WikiMQA use 32 samples per seed.  RQ2--RQ3
use Qwen2.5-1.5B-Instruct over eight update-condition families, three gaps, and
three seeds, yielding 72 profiles with 16 paired samples each.  Candidate
recovery windows are direct reuse, 1, 2, 4, 8, and 16 layers, and the complete
executable suffix.  The frozen condition set is a structured mechanism sweep
rather than a complete Cartesian product: it covers early key, value, query,
and MLP updates; middle query and MLP updates; and late key and MLP updates.
RQ4 applies 32 consecutive middle value-projection updates to
Qwen2.5-7B-Instruct.  Calibration uses two seeds, while held-out evaluation uses
three disjoint seeds, producing 96 held-out trajectories and 3,072 held-out
cache decisions per method.

KL is $D_{\mathrm{KL}}(p^{\star}\Vert p)$ at the first generated-token step;
top-1 agreement uses the same distributions, and task score evaluates the full
answer.  For RQ2, a candidate window is acceptable when KL is at most $0.05$,
top-1 agreement is at least $0.99$, and task-score drop from fresh is at most
$0.01$; the aggregate metrics must satisfy the same thresholds, with no paired
sample violating any threshold.  For RQ4, task score, KL, top-1 agreement, and
recomputation fraction are averaged over the 32 versions within each trajectory
and then across trajectories.  Cache-maintenance and total-request times are
instead summed over the 32 versions within each trajectory and then averaged
across trajectories; the main-paper maintenance-time column and request-time
comparison therefore report cumulative time per complete trajectory.
All RQ1 trajectories and sample-level outputs are included in the released
artifact; the main-paper figure reports their complete trends.

\subsection{Recomputation-Horizon Breakdown}

Table~\ref{tab:sup-horizon-target} decomposes the 72 profiles by update target
and position.  ``Direct'' denotes horizon zero, ``bounded'' a positive window
shorter than the complete executable suffix, and ``full'' the complete suffix.
Each row contains nine profiles from three seeds and three version gaps.

\begin{table}[t]
\centering
\small
\setlength{\tabcolsep}{3.5pt}
\begin{tabular*}{\columnwidth}{@{\extracolsep{\fill}}lrrl@{}}
\toprule
Condition & Suffix depth & D/B/F & Observed horizons \\
\midrule
Key--early & 28 & 2/6/1 & $\{0,1,8,16,28\}$ \\
Key--late & 1 & 7/0/2 & $\{0,1\}$ \\
Value--early & 28 & 3/2/4 & $\{0,1,4,28\}$ \\
Query--early & 28 & 6/1/2 & $\{0,16,28\}$ \\
Query--middle & 14 & 1/6/2 & $\{0,2,4,8,14\}$ \\
MLP--early & 28 & 3/1/5 & $\{0,16,28\}$ \\
MLP--middle & 14 & 6/2/1 & $\{0,2,8,14\}$ \\
MLP--late & 1 & 9/0/0 & $\{0\}$ \\
\midrule
All profiles & -- & 37/18/17 & $\{0,1,2,4,8,14,16,28\}$ \\
\bottomrule
\end{tabular*}
\caption{Recovery outcomes by updated module and position. D/B/F denotes
direct, bounded, and full-suffix outcomes.}
\label{tab:sup-horizon-target}
\end{table}

At gaps 1, 4, and 16, the direct/bounded/full splits are respectively
$14/5/5$, $13/7/4$, and $10/6/8$.  Both zero and full-suffix outcomes occur at
the same gap, confirming that version age alone is insufficient for recovery
planning.  For query and MLP updates, a one-layer candidate restarts execution
but replaces no affected KV state; it is therefore equivalent to stale reuse.
This equality holds for all corresponding rows in the frozen sample-level
analysis.

\subsection{Propagation-Proxy Correlations}

Table~\ref{tab:sup-tail-correlations} reports representative propagation
proxies associated with the unresolved suffix.  These quantities are distinct
from the output-space tail influence $T_b$ used by the theoretical certificate.
Within-condition correlations rank-center each seed--target--gap group,
removing between-condition scale effects.  Cumulative suffix measurements
preserve more signal than a single boundary measurement.

\begin{table}[t]
\centering
\small
\setlength{\tabcolsep}{4.5pt}
\begin{tabular}{lrr}
\toprule
Metric & Overall $\rho$ & Within $\rho$ \\
\midrule
Hidden tail sum & 0.873 & 0.690 \\
Hidden boundary & 0.666 & 0.320 \\
Key tail sum & 0.853 & 0.690 \\
Value tail sum & 0.853 & 0.690 \\
Final suffix-attention input & 0.892 & 0.620 \\
\bottomrule
\end{tabular}
\caption{Spearman correlation between candidate-window KL and representative
propagation metrics.}
\label{tab:sup-tail-correlations}
\end{table}

A fixed-effect log--log regression estimates an exponent of $2.14$ between KL
and final unrepaired suffix-attention input error, with within-condition
$R^2=0.115$.  Because suffix-attention input error is a propagation proxy rather
than the output-space $T_b$, the near-quadratic exponent is qualitatively aligned
with, but does not directly verify, the quadratic KL certificate.  Substantial
request-level variation remains unexplained.

\subsection{Taskwise RQ4 Results}

Table~\ref{tab:sup-rq4-taskwise} gives taskwise results for the four main
methods and two ablations.  Fixed-window recovery can obtain a slightly higher
mean task score than fresh execution on HotpotQA, but its KL and top-1
agreement show that this does not represent closer reproduction of the fresh
model.

\begin{table}[t]
\centering
\small
\setlength{\tabcolsep}{2.5pt}
\begin{tabular*}{\columnwidth}{@{\extracolsep{\fill}}lrrrr@{}}
\toprule
Method & Score & KL & Top-1 & Recomp. \\
\midrule
\multicolumn{5}{l}{\emph{HotpotQA}} \\
Fresh full & 9.41 & 0.0000 & 100.00 & 100.00 \\
Exact affected suffix & 9.41 & 0.0000 & 100.00 & 50.00 \\
Stale reuse & 7.95 & 0.3923 & 76.56 & 0.00 \\
\textsc{CacheReforge} & 9.13 & 0.0197 & 93.75 & 6.25 \\
Drift-only & 9.41 & 0.0595 & 89.19 & 3.12 \\
Fixed-window (2) & 9.99 & 0.1967 & 83.46 & 7.14 \\
\midrule
\multicolumn{5}{l}{\emph{2WikiMQA}} \\
Fresh full & 9.40 & 0.0000 & 100.00 & 100.00 \\
Exact affected suffix & 9.40 & 0.0000 & 100.00 & 50.00 \\
Stale reuse & 9.03 & 0.4847 & 92.38 & 0.00 \\
\textsc{CacheReforge} & 9.16 & 0.0473 & 94.34 & 4.63 \\
Drift-only & 8.99 & 0.0567 & 95.25 & 4.52 \\
Fixed-window (2) & 8.66 & 0.2430 & 92.64 & 7.14 \\
\bottomrule
\end{tabular*}
\caption{Taskwise held-out RQ4 results (all values except KL are percentages).
Each task contains 48 complete 32-version trajectories.}
\label{tab:sup-rq4-taskwise}
\end{table}

Exact affected-suffix recomputation matches fresh task and model-output
behavior.  Its quality and maintenance statistics are measured from 96 direct
32-version held-out trajectories (16 samples, three seeds, and two tasks),
covering 3,072 exact-suffix decisions.  These runs confirm zero KL, $100\%$
top-1 agreement, and $50\%$ recomputation.
Across a 32-version trajectory, \textsc{CacheReforge} averages 26.56 reuse,
2.94 bounded-recovery, and 2.50 exact-suffix actions.  Both bounded and exact
suffix actions use the same executable partial-recomputation path; their end
layers distinguish the two cases.

\FloatBarrier

\section{Reproducibility}

The reproducibility package is archived on Zenodo at
\href{https://doi.org/10.5281/zenodo.22951118}{doi:10.5281/zenodo.22951118}. It contains four RQ-specific packages with executable
source, configurations, raw outputs, regenerated analyses, paper aggregates,
and SHA-256 manifests. RQ4 additionally includes calibration runs, frozen risk
profiles, held-out trajectories, ablations, and per-sample provenance.
Machine-generated timestamps may remain in metadata.

The recorded environment is Linux on AArch64 with Python 3.12.13, PyTorch
2.7.1, torch-npu 2.7.1.post2, and Transformers 5.13.0, using bfloat16 on Ascend
NPUs.  The 7B isolated-gap runs use model sharding, whereas RQ4 uses
single-device deterministic execution.  Public configurations name the Qwen2.5
checkpoints and use relative data and output paths.

Development considered LoRA ranks $\{4,8,16\}$, update magnitudes from
$10^{-5}$ to $5\times10^{-2}$, and direct reuse, 1-, 2-, 4-, 8-, and 16-layer
windows plus the complete executable suffix.  Reported runs use rank 8;
mechanism studies use magnitude $10^{-3}$ and realistic QA/RQ4 use $10^{-2}$.
RQ4 uses a middle-layer value projection, yielding a 16-layer executable suffix
and allowing repeated partial recovery to be evaluated under a cache-producing
update.

RQ4 profiles were frozen on calibration seeds 53 and 61; seeds 7, 17, and 29
were held out.

\end{document}